\documentclass[runningheads,a4paper]{llncs}
\usepackage{amssymb}
\usepackage{graphicx}
\usepackage{siunitx}  
\usepackage{amsmath, amssymb} 
\usepackage{color}
\usepackage{textcomp}
\usepackage{adjustbox}
\usepackage{array}
\newcolumntype{L}[1]{>{\raggedright\let\newline\\\arraybackslash\hspace{0pt}}m{#1}}
\newcolumntype{C}[1]{>{\centering\let\newline\\\arraybackslash\hspace{0pt}}m{#1}}
\newcolumntype{R}[1]{>{\raggedleft\let\newline\\\arraybackslash\hspace{0pt}}m{#1}}
\newcommand{\noop}[1]{} 
\begin{document}
\frontmatter          
\pagestyle{empty}  
\mainmatter  

\title{Learning associations between clinical information and motion-based descriptors using a large scale MR-derived cardiac motion atlas}

\author{Esther Puyol-Ant\'on \inst{1} \and 
Bram Ruijsink \inst{1,2} \and
H\'el\`{e}ne Langet \inst{3} \and 
Mathieu De Craene \inst{3} \and
Paolo Piro \inst{4} \and
Julia A. Schnabel \inst{1} \and
Andrew P. King \inst{1}}

\authorrunning{E Puyol-Ant\'on et al.}   

\institute{School of Biomedical Engineering \& Imaging Sciences, King\textquotesingle s College London, UK. \and
St Thomas\textquotesingle{} Hospital NHS Foundation Trust, London, UK. \and
Philips Research, Medisys, Paris, France \and Sanofi R\&D, Paris}

\maketitle              
\begin{abstract} The availability of large scale databases containing imaging and non-imaging data, such as the UK Biobank, represents an opportunity to improve our understanding of healthy and diseased bodily function. Cardiac motion atlases provide a space of reference in which the motion fields of a cohort of subjects can be directly compared. In this work, a cardiac motion atlas is built from cine MR data from the UK Biobank ($\approx$ 6000 subjects). Two automated quality control strategies are proposed to reject subjects with insufficient image quality. Based on the atlas, three dimensionality reduction algorithms are evaluated to learn data-driven cardiac motion descriptors, and statistical methods used to study the association between these descriptors and non-imaging data. Results show a positive correlation between the atlas motion descriptors and body fat percentage, basal metabolic rate, hypertension, smoking status and alcohol intake frequency. The proposed method outperforms the ability to identify changes in cardiac function due to these known cardiovascular risk factors compared to ejection fraction, the most commonly used descriptor of cardiac function. In conclusion, this work represents a framework for further investigation of the factors influencing cardiac health.\\

\keywordname{ Cardiac motion atlas; non-imaging data; dimensionality reduction; multivariate statistics}
\end{abstract}
%
\section{Introduction}
Although much is known about the factors influencing healthy and diseased cardiac development, the recent  availability of large scale databases such as the UK Biobank represents an excellent opportunity to extend this knowledge. The wide range of clinical information contained in such databases can be used to learn new associations with indicators of cardiac function. In this paper we propose a framework for performing such an investigation. \\
Left-ventricular (LV) function has been traditionally assessed using global indicators such as ejection fraction and stroke volume. The main drawback of such indicators is that they do not provide localised information about ventricular function, and so their diagnostic power is limited. Motion parameters such as displacements allow for a more spatially and temporally localised assessment of LV function.  Statistical cardiac motion atlases have been proposed as a way of making direct comparisons between motion parameters from multiple subjects \cite{bai2015bi,peressutti2017framework}. We use the UK Biobank database to build a cardiac motion atlas from cine MR images using around 6000 subjects with no known cardiovascular disease. To ensure that the estimated motion parameters are accurate, two automated quality control strategies are proposed to reject subjects with insufficient image quality. We use motion descriptors extracted from the atlas to identify relationships between cardiac function and other clinical information.\\
Previously, several papers have tried to analyse the influence of clinical information such as smoking, alcohol and body weight on cardiovascular disease \cite{jitnarin2014relationship,mukamal2006alcohol,nadruz2016smoking}. However, these studies were based on a limited number of factors and used relatively crude indicators of cardiac health, such as categorical variables indicating disease state. In contrast, we exploit the large scale atlas to make use of more localised motion parameters and also examine a wider range of clinical information. \\
\textbf{Contributions:} Our scientific contributions in this paper are threefold. First, we present, for the first time, a large scale cardiac motion atlas built from the UK Biobank database. Second, we propose an automated quality control approach to ensure the robustness of the extracted motion descriptors. Third, we use the atlas to learn associations between descriptors of cardiac function and non-imaging clinical information. This automatic framework represents an ideal tool to further explore factors influencing cardiac health.
\section{Materials}
\label{sec:materials}
The UK Biobank data set contains multiple imaging and non-imaging information from more than half a million 40-69 year-olds. From this data set, only participants with MR imaging data were considered for analysis. Furthermore, participants with non-Caucasian ethnicity, known cardiovascular disease, respiratory disease, diabetes, hyperlipidaemia, haematological disease, renal disease, rheumatological disease, malignancy, symptoms of chest pain or dyspnoea were excluded. \\
MR imaging was carried out on a 1.5 Tesla scanner (Siemens Healthcare, Erlangen, Germany) \cite{petersen2016uk}. Short-axis (SA) stacks covering the full heart, and two orthogonal long-axis (LA) planes (2-chamber (2Ch) and 4-chamber (4Ch) views) were available for each subject (TR/TE = \SI{2.6/1.10}{\milli\second}, flip angle = \ang{80}). In-plane resolution of the SA stack and LA images was \SI{1.8}{\milli\metre}, with slice thickness of  \SI{8}{\milli\metre} and \SI{6}{\milli\metre} for SA and LA respectively. 50 frames were acquired per cardiac cycle  (temporal resolution $\approx$14-24 ms/frame). In addition, the following non-imaging parameters were used for the statistical analysis: age, smoking status (smoker/non-smoker), body mass index (BMI), body fat percentage (BFP), basal metabolic rate (BMR), hypertension and alcohol intake frequency (occasionally/regularly). Demographics of the analysed cohort are presented in Table \ref{tab:demographics}.
\begin{table}
\centering 
\caption{Baseline characteristics for the healthy cohort from the UK Biobank data set. All continuous values are reported as mean (SD), while categorical data are reported as number (percentage).}
\begin{tabular}{l c } 
& Total \\ \hline
Number of participants & 6002\\
Male gender (n(\%))& 3046 (50.8\%)\\
Age(years) & 61 (8)\\
Body mass index (BMI) ($kg/m^2$) & 26.4 (4.2)\\
Body fat percentage (BFP) ($\%$) & 28.2 (7.7) \\
Basal metabolic rate (BMR) ($KJ$) & 6650.1 (1324.9))\\
Smokers (n(\%)) & 1011 (38.2\%) \\
Regular alcohol intake (n(\%))& 1407 (48.9\%) \\  
Hypertension (n(\%)) & 2903 (48.4\%)\\
Ejection fraction (EF)(\%) & 59.2 (6.3\%)\\
\end{tabular}
\label{tab:demographics}
\end{table}
\section{Methods}
\label{sec:methods}
In the following Sections we describe the estimation and analysis of the atlas-based motion descriptors. Details of the atlas formation procedure are reported in Sect. \ref{subsec:atlas}. Sect. \ref{sbsc:QC} details the quality control (QC) strategies proposed.  Sect. \ref{subsec:dimensionality_reduction} describes the different dimensionality reduction algorithms used to characterise cardiac function. Sect. \ref{subsec:statistical_analysis} presents the statistical methods used to identify relationships between the motion descriptors and non-motion clinical information.

\subsection{Motion Atlas Formation}
\label{subsec:atlas}
This section outlines the procedure used to build the motion atlas, which is based on the work published in \cite{bai2015bi} and \cite{peressutti2017framework}.\\
\indent \textbf{LV Geometry Definition.} A fully-convolutional network (FCN) with a 17 convolutional layer VGG-like architecture was used for automatic segmentation of the LV myocardium and blood-pool at end-diastole (ED) for SA and LA slices \cite{bai2017semi,sinclair2017fully}. Each convolutional layer of the network is followed by batch normalisation and ReLU, except the last one, which is followed by the softmax function. In the case of the SA stack, each slice is segmented independently, i.e. in 2D. From the segmentations, a bounding box was generated and used to crop the image to only include the desired field of view, improving pipeline speed and reducing errors in motion tracking. The MR SA and LA segmentations were used to correct for breath-hold induced motion artefacts using the iterative registration algorithm proposed in \cite{sinclair2017fully}. The motion-corrected LA/SA segmentations were fused to form a 3D smooth myocardial mask. An open-source statistical shape model (SSM) was optimised to fit to this LV binary mask following an initial rigid registration using anatomical landmarks \cite{bai2015bi}. The use of landmarks ensured that the LV mesh was aligned to the same anatomical features for each patient, and the endocardial and epicardial surfaces of the mesh were smoothly fitted to the myocardial mask, providing surface meshes with point correspondence for each of the subjects and a 17 AHA regions segmentation. \\
\textbf{Motion Tracking}. A 3D GPU-based B-spline free-form deformation (FFD) was used \cite{rueckert1999nonrigid} to estimate LV motion between consecutive frames. Subsequently, the inter-frame transformations were estimated and then composed using a $3D+t$ B-spline to estimate a full cycle  $3D+t$ transformation.\\
\textbf{Spatial Normalisation}. Similar to \cite{peressutti2017framework}, we transform each mesh to an unbiased atlas coordinate system using a combination of Procrustes alignment and Thin Plate Spline transformation. We denote the transformation for each subject $n$ from its subject-specific coordinate system to the atlas by  $\phi_n$.\\
\textbf{Medial Surface Generation.} A medial surface mesh with regularly sampled vertices ($\approx 1000$) was generated from the SSM epicardial and endocardial meshes using ray-casting and homogeneous downsampling followed by cell subdivision. \\
\textbf{Motion Reorientation.} The displacements ($\boldsymbol{u}_n$) for each subject $n$ were reoriented into the atlas coordinate system under a small deformation assumption using a push-forward action: $\boldsymbol{u}^{atlas}_{n} = \boldsymbol{J}{(\phi_n(\boldsymbol{r}_{n}))} \boldsymbol{u}_{n}$ \cite{peressutti2017framework}, where $\boldsymbol{J}{(\cdot)}$ refers to the Jacobian matrix, and $\boldsymbol{r}_{n}$ are the positions of the vertices in the ED frame. \\
\textbf{Transform to Local Coordinate System.} For each subject $n$ the displacements in the atlas coordinate system $\boldsymbol{u}^{atlas}_n$  were projected onto a local cylindrical coordinate system $\boldsymbol{x}^{atlas}_n$, providing radial, longitudinal, and circumferential information. The long axis of the LV ED atlas medial mesh was used as the longitudinal direction.
\subsection{Quality control}
\label{sbsc:QC}
To ensure that the atlas was formed only from subjects with high quality cine MR images and robust motion descriptors, two QC methods were implemented to automatically reject subjects with insufficient quality or incorrect segmentations and/or motion tracking. These two methods are:\\
\textbf{QC1.} SA images with insufficient image quality were excluded using two deep learning based approaches. First the 3D convolutional neural network (CNN) described in  \cite{oksuz2018miccai} was used to automatically identify images with motion-related artefacts such as mistriggering, arrhythmia and breathing artefacts. Second, subjects with incorrectly planned 4Ch LA images were automatically excluded using the CNN described in \cite{oksuz2018isbi}. This pipeline identified images containing the left ventricular outflow tract, which is a common feature of poorly planned 4Ch images. A total of 806 subjects were rejected using the QC1 step.\\
\textbf{QC2.} Our second QC measure aimed to detect segmentation and/or motion tracking errors. LV volume curves over the cardiac cycle were computed based on the LV geometry and motion fields. Based on these curves, an automatic pipeline was used to identify curves with unrealistic properties. The LV curve was divided into diastole and systole based on the ED and end-systolic (ED) points, where ED corresponds to the point with maximum volume, and ES corresponds to the point with minimum volume. These two curves were treated independently. In the systole part of the curve, the downslope should be smooth and continuous and therefore the first derivative should only contain one peak. Cases with more than one (automatically identified using the first and the second derivative of the LV curve) peak were therefore excluded. In the diastole part the curves were divided into rapid inflow, diastasis and atrial systole phases. In the rapid inflow phase the upslope should be smooth and continuous and the first derivative should only contain one peak. Cases with more than one peak were therefore excluded. In the diastasis phase the curve should be flat and with no variation, so the first and second derivatives were used to exclude any cases with changes of volume higher than 10\%. Finally, in the atrial systole phase the upslope should be smooth and continuous and the first derivative should only contain one peak. Cases with more than one peak were therefore excluded. The QC2 step excluded a total of 64 cases.  
\subsection{Dimensionality reduction}
\label{subsec:dimensionality_reduction}
In this work, we propose to learn descriptors of cardiac function from the motion atlas using dimensionality reduction techniques. More specifically, three techniques were compared: Principal Component Analysis (PCA) \cite{hotelling1933analysis}, Locally Linear Embedding (LLE) \cite{Roweis2323}, and stacked denoising autoencoders (SDA) \cite{vincent2010stacked}. To form the inputs to these three algorithms, the local displacements per AHA region were averaged and these were concatenated into a column vector such that for subject $n$, $\boldsymbol{x}^{atlas} \in  \mathbb{R}^L$, where $L= 3AT$, $T$ is the number of cardiac phases and $A$ is the number of AHA regions ($17$) in the atlas medial surface mesh. The column vectors for each subject were combined to produce a matrix $X=[\boldsymbol{x}^{atlas}_1,...,\boldsymbol{x}^{atlas}_N]$, where $N$ is the number of subjects. The matrix $X$ is the input for all of the proposed dimensionality reduction techniques, and the output of all of them is a matrix $D \in \mathbb{R}^{(d \times N)}$, where $d$ is the size of the low dimensional space.
\subsection{Statistical analysis}
\label{subsec:statistical_analysis}
Multivariate statistics were used to find relationships between the low-dimensional motion descriptors and non-imaging clinical information. For categorical variables (i.e. smoking status, hypertension and alcohol intake) a multivariate Hotelling\textquotesingle{}s $T^2$ test (assuming unequal variance) \cite{hotelling1992generalization} was used to test for differences between groups, and the $p$-value was recorded. For continuous variables (age, BMI, BFP, BMR) a multivariate regression was used to compute the $R$ statistic that measures the degree of relationship between the data. The $R$ statistic and associated $p$-values were recorded. All the recorded $p$-values were adjusted using Bonferroni\textquotesingle s correction for multiple tests.  \\
Furthermore, we analysed the impact of using a cardiac motion atlas compared to standard clinical global measures such as ejection fraction (EF), the current most used clinical global measure of cardiac function. To this end, the EF for each subject was computed from the tracked meshes. For categorical variables, similar to the previous case, a paired $t$-test (assuming unequal variance) was used to test for differences between groups, and the $p$-value was recorded. For continuous variables a linear regression was used to compute the $R$ statistic and the associated $p$-value. As before, all the recorded $p$-values were adjusted using Bonferroni\textquotesingle s correction for multiple tests. 
\section{Experiments and Results}
\label{sec:results} 
For all of the proposed dimensionality reduction algorithms the number of dimensions $d$ retained was optimised. In the case of PCA, $d$ was selected to retain 99\% of the variance; in the case of LLE a grid search algorithm was used to vary $d$ between 2 and 256 and the number of neighbours $s$ was fixed to 10. Then, $d$ and $s$ were selected to minimise the reconstruction error $\epsilon$. Finally, SDA was applied with 5 layers with size $[L, 2000, 1000, 500, d]$, a learning rate of 0.001, a corruption level of 50\%, and number of epochs 500. The value of $d$ was optimised using a grid search between 2 and 256, and $d$ was selected to minimise the reconstruction error $\epsilon$. For the three dimensionality reduction algorithms, the optimised $d$ were: $d_{PCA}$ = 932; $d_{LLE}$ = 32 and $\epsilon_{LLE}$ = 0.37 mm; $d_{SDA}$ = 50 and $\epsilon_{SDE}$ = 0.15mm.\\

Table \ref{tab:results} shows the computed $p$-values with Bonferroni\textquotesingle s correction and $R$ statistics for the different clinical data for the different dimensionality reduction algorithms, and also using only EF as a motion descriptor. Bold numbers show $p$-values less than 0.05 (i.e. that showed statistical significance at 95\% confidence). The results show that for all of the dimensionality reduction methods there was no statistical significance for age and BMI, while BFP, BMR, smoking status, hypertension and alcohol consumption are shown to be associated with cardiac function. Furthermore, the R statistics for BFP and BMR for all of the non-linear dimensionality reduction algorithms are around 60\%, which indicates a clear linear relationship with the data. Using only EF we can see that there is no statistical significance with any of the clinical information, and the $R$ values are lower compared to the use of the cardiac motion atlas. Non-linear dimensionality reduction algorithms, i.e. LLE and SDA show lower $p$-values and higher $R$ statistics compared to linear PCA for all of the clinical information analysed. For BFP, BMR, smoking status, hypertension and alcohol consumption, LLE shows a better linear fitting, with slightly higher $R$ statistics compared to SDA. However, the difference between the two methods is not significant. \\
\begin{table}[h]
\centering
\begin{tabular}{ L{19mm} C{12mm} C{12mm} C{12mm} C{12mm} C{12mm} C{12mm} C{12mm} C{12mm}}
\hline
\multicolumn{1}{ l }{}  & \multicolumn{4}{ c }{\textbf{$R$ statistics (\%)}} & \multicolumn{4}{ c }{\textbf{$p$-value}}\\ \hline
\multicolumn{1}{ l }{}  & PCA & LLE & SDA & EF & PCA & LLE & SDA & EF\\ \hline
Age & 48.45 & 49.03 & 56.11 & 39.67 & 0.31 & 0.29 & 0.28 & 0.38 \\
BMI & 45.19 & 57.47 & 59.83 & 25.87 & 0.20 & 0.16 & 0.18 & 0.31 \\
BFP & 48.22 & 64.20 & 64.63 & 23.98 & \textbf{0.05} & \textbf{0.03} & \textbf{0.05} & 0.68 \\
BMR & 55.67 & 64.60 & 55.84 & 21.96 & \textbf{0.05} & \textbf{0.01} & \textbf{0.04} & 0.21 \\
Smoking   & & & & & \textbf{0.05} & \textbf{0.02} & \textbf{0.04} & 0.56 \\
Alcohol 		 & & & & & \textbf{0.05} & \textbf{0.03} & \textbf{0.02} & 0.62 \\
Hypertension & & & & & \textbf{0.03} & \textbf{0.01} & \textbf{0.03} & 0.15  \\
\end{tabular}
\caption{Association of clinical information with estimated motion descriptors from the atlas represented by computed $p$-values and $R$ statistics using multivariate statistics. The $p$-values and $R$ statistics for EF were computed using paired t-tests and linear regression respectively. Bold numbers show $p$-values after Bonferroni's correction less than 0.05, which are considered to be statistically significant at 95\% confidence.}
\label{tab:results}
\end{table}
\newpage
\section{Discussion}
\label{sec:discussion}
In this work, we have demonstrated the derivation of motion atlas-based descriptors of cardiac function and investigated their association with a range of clinical information. This is the first time that a cardiac motion atlas has been formed from a large number of subjects from the UK Biobank, and this has enabled us to produce a framework for investigating factors influencing cardiac health. We show that our method was able to identify factors known to impact cardiac health (directly or indirectly), such as hypertension, smoking, alcohol intake and body composition. Contrary to previous smaller studies, age was not related to large changes in cardiac function. We were unable to assess whether the apparent lack of association was influenced by a selection bias due to anticipated better survival of healthy versus relatively unhealthy individuals, the relatively small spread in ages (45-70) compared to other studies, the clustering of most participants in the middle age range (55-65), or was a consequence of the dimensionality reduction algorithms. We will investigate these possibilities in future work. In addition, we demonstrated that the use of a cardiac motion atlas has added value over using only global indicators such as EF. The proposed method represents an important contribution to furthering our understanding of the influences on cardiac function. We used the selected categories as an illustration of the capabilities of the atlas method. However, the univariate analysis that we performed does not capture the complex interrelation of health factors, age and cardiac function. Future work will focus on a deeper analysis of the impact of these parameters and will also include further clinical characteristics. In this work, displacements were used as a representation of cardiac motion. However, other representations such as velocity and strain can characterise motion in different ways. Our future work also includes extension of the motion atlas to incorporate strain parameters. In conclusion, this work represents an important contribution to furthering our understanding of the influences on cardiac function and opens up the possibility of learning the relationships between motion descriptors and further non-imaging information.

\section*{Acknowledgements}
This work is funded by the Kings College London \& Imperial College London EPSRC Centre for Doctoral Training in Medical Imaging (EP/L015226/1) and supported by the Wellcome EPSRC Centre for Medical Engineering at Kings College London (WT 203148/Z/16/Z). This research has been conducted using the UK Biobank Resource under Application Number 17806.

\bibliographystyle{splncs03}
\bibliography{mybibfile}
\end{document}